%% file: sample-sigconf.tex
\documentclass[sigconf]{acmart}

\usepackage{booktabs} 
\usepackage{graphicx}

\setcopyright{acmcopyright}

\acmDOI{xx.xxx/xxx_x}

\acmISBN{978-1-4503-9517-5/23/03}

\acmConference[SAC'23]{ACM SAC Conference}{March 27 –April 2, 2023}{Tallinn, Estonia}
\acmYear{2023}
\copyrightyear{2023}

\acmArticle{4}
\acmPrice{15.00}

\begin{document}
\title{Ensemble Creation via Anchored Regularization for Unsupervised Aspect Extraction}
  

\author{Pulah Dhandekar}
\affiliation{%
  \institution{Applied Research, Thoucentric}
  \city{Bangalore} 
  \country{India}
}
\email{pulahdhandekar@thoucentric.com}

\author{Manu Joseph}
\affiliation{%
  \institution{Applied Research, Thoucentric}
  \city{Bangalore} 
  \country{India}
}
\email{manujoseph@thoucentric.com}

\begin{abstract}

Aspect Based Sentiment Analysis is the most granular form of sentiment analysis that can be performed on the documents / sentences. Besides delivering the most insights at a finer grain, it also poses equally daunting challenges. One of them being the shortage of labelled data. To bring in value right out of the box for the text data being generated at a very fast pace in today’s world, unsupervised aspect-based sentiment analysis allows us to generate insights without investing time or money in generating labels. From topic modelling approaches to recent deep learning-based aspect extraction models, this domain has seen a lot of development. One of the models that we improve upon is ABAE that reconstructs the sentences as a linear combination of aspect terms present in it, In this research we explore how we can use information from another unsupervised model to regularize ABAE, leading to better performance. We contrast it with baseline rule based ensemble and show that the ensemble methods work better than the individual models and the regularization based ensemble performs better than the rule-based one.
\end{abstract}




\maketitle

\input{samplebody-conf}

\bibliographystyle{ACM-Reference-Format}
\bibliography{sample-bibliography} 

\end{document}

%% file: samplebody-conf.tex
\section{Introduction}


Sentiment analysis as a domain of Natural Language Processing has been studied for a lot of years. According to a survey conducted by \cite{Wankhade2022}, Sentiment analysis has applications in Business analysis, Healthcare and Medical domain, Review analysis, Stock Market, Voice of the Customers and Social Media Monitoring. It can be used to make product enhancements and develop marketing strategies to target relevant audience. It can be used for market research and competitor analysis. Reputation management is another key area where sentiment analysis can help. In such areas, a much more targeted sentiment as provided by Aspect Based Sentiment Analysis (ABSA) can prove to be more insightful. Out of various levels at which the sentiment can be extracted, this research mainly focuses on the subtask of Aspect Extraction from the task of Aspect level sentiment analysis. Instead of simply assigning the polarity at a sentence / corpus level, the ABSA task focuses on extraction of the aspects in the sentence and finding the polarity for each of the aspects in the sentences. Though it is possible to have multi-aspect sentences, this research revolves mainly around the single aspect sentences due to our choice of models. For sentences like "The food is great" the Aspect extraction task would involve identification of aspect "food" from the sentence. There have been many works which tackle ABSA as a supervised problem. But because of lack of labelled data in real-world datasets, unsupervised approaches are attractive.

The aspect extraction task aims to ultimately assist in Aspect level sentiment analysis. The major contribution of the research are:
\begin{enumerate}
    \item We propose an innovative ensemble of unsupervised aspect extraction models using anchored regularization
    \item The novel method allows us to couple any model with Attention Based Aspect Extraction (ABAE) model as long as we are able to extract Aspect Categories from the prior model in the pipeline.
    \item It overcomes the need of explicit rule creation as done in the Rule Based Ensemble Creation, thus saving valuable time and efforts.
\end{enumerate}

\section{Literature Review}

The Unsupervised Aspect Extraction process has seen a lot of development in the recent years. Inspired by the idea of LDA, one of the first Aspect extraction model was Local-LDA \cite{Brody2010}. This model, treated each sentence as a separate document, which was followed by a clustering algorithm to extract the aspects.A few years later, another model called BTM \cite{Yan2013} was proposed which creates a generative Biterm model. It explicitly models word co-occurrence for topic learning. Rich global word co-occurrence patterns are used for better revealing topics. Biterm is an unordered word pair that co-occurred in a short document. 

With the advent of the Deep learning based models a significant improvement in model performance was observed. ABAE \cite{He2017} is one of the first Autoencoder like attention based models to be able to extract the Aspect terms from the corpus. Using the CitySearch and BeerCorpus it is aims to recreate the original sentence and generate an aspect terms embeddings such that they are the most dissimilar to each other, hereby decreasing the possibility of the embeddings being close to each other in the embedding space. Similar to the ABAE model, AE-SA and AE-CSA \cite{Luo2019} use the Encoder - Decoder based model that leverage the concept of hierarchical attention and context enhanced attention to extract aspects from the sentences in an unsupervised manner. 

The Contrastive Attention (CAt) \cite{Tulkens2020} algorithm although not a deep learning based algorithm, uses the concept of attention which has its roots in deep learning. It is a novel single-head attention mechanism which is based on the RBF kernel. This method is found to have given a considerable boost in the model performance and has also made the model more interpret-able while working in limited scope. In addition to this, the model only uses a POS tagger and word embeddings, which enables the user to apply it directly to new domains and a variety of languages. In comparison to its deep learning counterparts, it is a considerably light-weight algorithm that is easy to implement and fast to run. 

Along with the gains made in terms of simplicity, some researchers stuck to the main idea of deep learning and made advancement by looking at the nearby and overall context together. One such algorithm called LCC+GBC \cite{Liao2020} model uses sequence of words in sentence to create the local context. The Bag of words (BOW) representation of a sentence is used to create the global context. Local context is captured with the use of LSTMs and the global context is captured with the help of an encoder which generated the global representation. These are both combined with the help of attention mechanism. For decoding, the global representation and globally scoped local representation are used together along with a SoftMax activation function. One of the models that tried to build on top of the learnings by ABAE model is the CMAM \cite{Sokhin2020} model that introduces a novel convolutional multi-attention mechanism which helps us find the aspects in the form of (aspect, term) pairs. It can perform better than the ABAE model in terms of F1 score for some of the aspects. 

In addition to the above methods there also exist hybrid methods \cite{SinghChauhan2020} which first create labelled through rules and then use it for training. For instance, this method first uses the lexical and syntactical method to extract aspects and then uses the results as labels for the attention model. The proposed model is a two-step mixed unsupervised model which combines linguistic patterns with deep learning techniques for the improvement of Aspect Term Extraction task. First, rule-based methods are used to extract the single word and multi-word aspects, which further pruned by domain-specific relevant aspects using fine-tuned word embedding. The next step is to use the extracted aspects in the first step which are used as label data to train the attention-based deep learning model for aspect-term extraction. 

Some of the research in aspect extraction came as a part of the solutions that aimed to solve the problem of Unsupervised ABSA as a whole. One such method proposed is a Restricted Boltzmann machine-based method that utilizes them and have proposed the SERBM (Sentiment Extraction Restricted Boltzmann Machine) model \cite{Wang2015} where instead of the hidden nodes being homogeneous, they are heterogeneous such that, a set of hidden nodes correspond to aspects, a set for sentiments and a set for background. It uses priors based on aspects detected via LDA model and sentiments detected via SentiWordNet \cite{Baccianella2010} to regularize the RBM loss function. Another such method proposes AutoPhrase algorithm \cite{Giannakopoulos2017} where quality phrases are extracted using syntactical rules and sentiment lexicon is used on order to find the sentiment of the aspect terms. The quality phrases are then pruned as per some threshold for the dataset, followed by rules such as it shouldn’t be a stop word, and should be present in pruned quality phrases. After applying some POS rules and domain specific rules, the automatically labelled dataset is created which is then used to train Bi-LSTM and CRF models.

\section{Research Methodology}

    

\subsection{ABAE model}
\begin{figure}[h]
\centering
\includegraphics[width=0.45\textwidth]{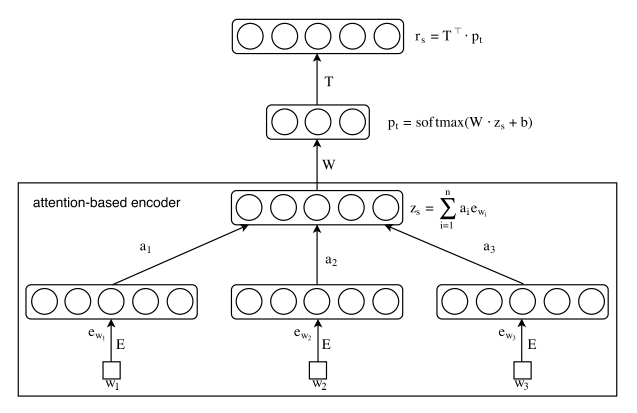}
\caption{ABAE model architecture \cite{He2017}}
\label{fig:abae}
\end{figure}

Attention Based Aspect Extraction (ABAE) \cite{He2017} is the first Unsupervised Aspect Term Extraction model to use deep learning and sentence recreation to extract aspects from the sentences without the use of any pseudo labelled data. Figure \ref{fig:abae} shows a high-level architecture of ABAE model.  Given embedding matrix $E \in R^{\{ V x d \}} $ with the vocabulary of size $V$ and an embedding size of $d$, the goal of the method is to learn an Aspect Term matrix $T \in R^{\{ k x d \}} $ where $k$ denote the predefined numbers of Aspect Categories and $k << V$. Here, each aspect term vector in matrix $T$, shares the same embedding space as the words in vocabulary $V$.

\begin{equation} \label{di}
d_{i} = e_{w_{i}}^T.M.y_{s}
\end{equation}

\begin{equation} \label{ai}
a_{i} = \frac{exp(d_{i})}{\sum_{j=1}^{n}exp(d_{j})}
\end{equation}

\begin{equation} \label{zs}
z_{s} = \sum_{i=1}^{n} a_{i} e_{w_{i}}
\end{equation}

\begin{equation} \label{pt}
p_{t} = softmax(W.z_{s} + b)
\end{equation}

\begin{equation} \label{rs}
r_{s} = T^T.p_{t}
\end{equation}

Equations \ref{di} to \ref{rs} denote the process of  obtaining the reconstructed sentence $r_{s}$ which is a linear combination of aspect embeddings. The aspect embeddings are learnt in matrix $T$, which is multiplied by the attention weights $p_{t}$ to obtain $r_{s}$. $z_{s}$ is the representation of the sentence in the form of dot product between the attention weights $a_{i}$ and each of the word in the sentence $e_{w_{i}}$. The parameters $ \theta = (T , M, W , b) $ are learnt as a part of the model training and $E$ is learnt beforehand through the training of Word2Vec model.


\begin{equation} \label{abae_og}
L(\theta)_{abae} = J(\theta) + \lambda U(\theta)
\end{equation}

\begin{equation} \label{Jtheta}
J(\theta) = \sum_{s \in D}\sum_{i=1}^{m}max(0,1-r_{s}z_{s}+r_{s}n_{i})
\end{equation}

\begin{equation} \label{Utheta}
U(\theta) = \| T_{n}.T_{n}^T - I \|
\end{equation}

The loss function of ABAE as seen in Equation \ref{abae_og} consists of \ref{Jtheta} and \ref{Utheta} terms that are collectively minimized. The term $J(\theta)$ as defined in Equation \ref{Jtheta} corresponds to the Max Margin loss which tries to maximize the similarity of reconstructed sentence with the true samples and minimize the similarity with the negative samples. A negative sample is any sentence randomly chosen other than the current sentence. This forces the model to not overfit on the current sentence. Maximizing similarity with true samples includes maximizing the product $r_{s}.z_{s}$ as this would make the model learn parameters such that the reconstructed sentence representation and weighted sentence representation are as close to each other as possible. It also tries to minimize $r_{s}.n_{i}$ where $n_{i}$ is nothing but the sentence embedding of a negative samples. In addition to this, the loss function also tries to minimize the dot product of aspect embedding matrix \textbf{T} with the transpose of itself. This forces the aspects to be orthogonal to each other and enforces diversity in the detected aspects. The term $U(\theta)$ as defined in Equation \ref{Utheta} refers to this orthogonality constraint used as regularization.

In this research two kinds of ensembles are explored namely The Rule Based Ensemble and The Regularization based ensemble. The setup for both these type of ensembles is discussed in this section. In case of the rule based ensemble, Part of Speech (POS) Tags are used to identify the nouns and the adjectives in the sentences to identify the aspects. In case of Regularization based ensemble, aspects identified by the prior model are fed to the Attention Based Aspect Extraction (ABAE) model as a regularization term in the loss function.

\subsection{Rule Based Ensemble}
The Rule Based Ensemble creation can be understood via the algorithm mentioned in this section. A point to note here is that, since CAt model can only perform well on the subset of aspect categories, this disambiguation logic is only used for those subsets. The predictions for the rest of the categories comes from either ABAE model or "miscellaneous" prediction. The algorithm is as follows:
\begin{enumerate}
    \item Obtain POS tags from each of the sentences using out of the box POS taggers available in spacy.
    \item Extract Nouns and Adjectives from the sentences and create a noun list or a combined list of nouns and adjectives. We call this candidate list.
    \item Obtain predictions from the individual models for each of the sentences.
    \item Assign labels as is to the sentences where all models mutually agree upon the predictions.
    \item For sentences with conflicting predictions coming from the models:
    \begin{enumerate}
        \item If candidate-list is not empty:
        \begin{enumerate}
            \item Calculate cosine similarity between each element of candidate list and aspect categories.
            \item Assign aspect of the highest similarity pair as prediction.
        \end{enumerate}
        \item If candidate-list is empty and if the prediction is unassigned use miscellaneous.
        \item For any unassigned predictions hereafter, prioritize ABAE prediction and assign 'miscellaneous' as the category for sentences missed by ABAE.
    \end{enumerate}
\end{enumerate}


\subsection{Regularization Based Ensemble}
The main idea behind the Regularization based ensemble is to first predict the aspect labels through the prior model in the pipeline and feed it to the loss function of the latter model such that, the regularization term penalizes the model to a greater extent if the aspect detected by the latter model are very different from those detected by the prior model. This allows the latter model (ABAE in this case) to obtain information about how the aspects for the sentence might look like which contributes to the overall learning process. Thus, we anchor the predictions generated by the latter model on the ones generated by the prior model. This concept can be applied to the coupling of any other model with ABAE and like models.To achieve this, an additional regularization term $K(\theta)$ is added to the existing loss function as seen in Equation \ref{abae_new}. 

\begin{equation} \label{abae_new}
L(\theta) = J(\theta) + \lambda U(\theta) + \sigma K(\theta)
\end{equation}

\begin{equation} \label{full_dot}
G = r_{s}.E_{emb}
\end{equation}

\begin{equation} \label{rel_dot}
G_{rel} = G \circ I
\end{equation}

\begin{equation} \label{min_eq}
K(\theta) = || G_{rel} - I ||_{2}
\end{equation}

Equation \ref{abae_og} refers to the original loss function used by ABAE model. To this loss function, we add the term $\sigma K(\theta)$ as in equation \ref{abae_new}. $r_{s}$ is the normalized reconstruction of sentences using aspect embedding matrix and attention in ABAE as calculated in equation \ref{rs}. As a result of this operation, each vector in the matrix $r_{s}$ is a representation of the aspect that best summarizes the original sentence. $E_{emb}$ is the embedding of the label extracted from the prior model for all sentences, represented in the space of ABAE model, and has also been normalized. In this normalization, each vector is divided by the L2-norm of itself. It shares the same vector space as the embeddings matrix $E$. $\sigma$ is a hyper parameter that controls the strength of regularization provided by the term. 

The result of equation \ref{full_dot} consists of dot product of all sentence reconstructions with all prior model label embeddings. The reconstruction matrix $r_s$ is the matrix of the best representation of the sentence by an aspect, picked by ABAE for each of the sentences respectively. $E_{emb}$ is the matrix of labels picked by the prior model for each of the sentences. The dot product of both these matrices, results in $G$ where the $ij^{th}$ entry corresponds to the dot product of the aspect approximation picked by ABAE to represent the $i^{th}$ sentence and the label picked by the prior model for the $j^{th}$ sentence. But, we are only concerned with sentence reconstructions and prior model label embeddings of corresponding sentences i.e. the aspect approximation of the $i^{th}$ sentence in $r_{s}$ and the $i^{th}$ prior label in $E_{emb}$. These embeddings are present only in the diagonal of the matrix $G$. To retain only these dot products and to set the rest as zero, we perform an element-wise multiplication of $G$ with identity matrix $I$. It makes sure that we only consider the correct (diagonal) elements for regularization. This gives us the term $G_{rel}$ as in equation \ref{rel_dot}. The elements in the diagonal of the matrix $G_{rel}$ range from -1 to 1 as the dot products are between normalized vectors. As a result, if we subtract an Identity matrix from $G_{rel}$ and square the elements, we get the maximum value for an element as 4 when the respective dot product is -1. This corresponds to the corresponding vectors of $r_{s}$ and $E_{emb}$ being the most different. The minimum value is attained, when the dot product is 1 and the respective value is 0. This corresponds to the vectors being exactly the same. This is the operation that takes place that results in $K(\theta)$ as in equation \ref{min_eq}. It is nothing but the summation of squared difference between the elements of $G_{rel}$ and $I$. Overall it tells us, how different are the reconstructed sentences and the prior labels from each other. This difference, when calculated over all sentences is the penalty incurred by the model.

\begin{figure}[h]
\centering
\includegraphics[width=4cm, height=13cm]{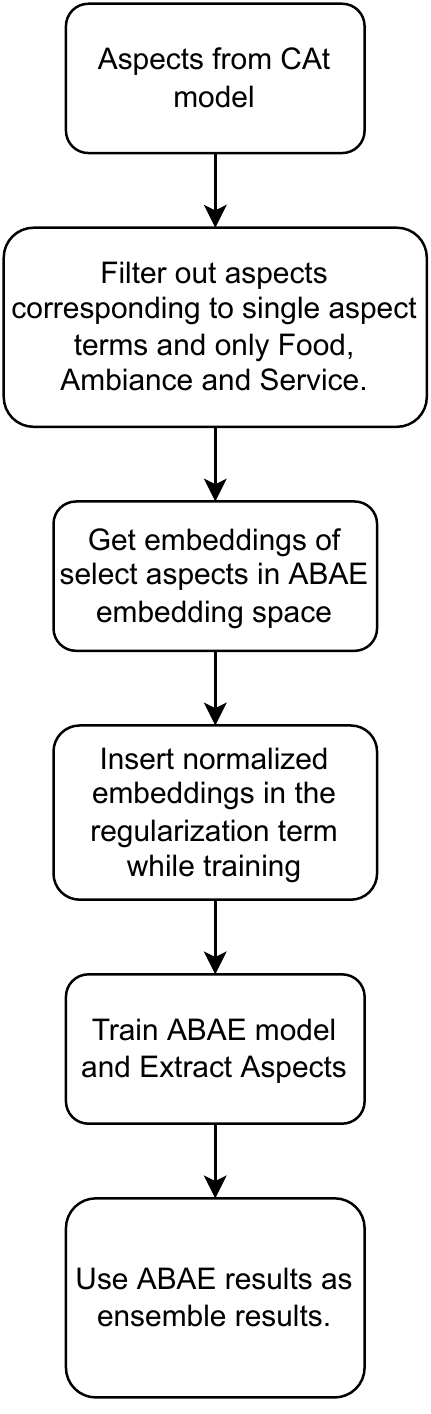}
\caption{Regularization Based Ensemble Representation}
\end{figure}


\section{Results and Discussions}
The datasets used for this research were CitySearch Corpus data set and Semeval 2014 data set \cite{Pontiki2014}. To make sure that the evaluation dataset has the labels present, we used the entirety of CitySearch Corpus data set along with unlabelled subset of Semeval 2014 dataset for training and used the labelled portion of Semeval 2014 dataset for evaluation. Out of all sentences, we filter out only the single aspect sentences since the models in the pipeline are capable of dealing with only single aspect sentences. Since the prior (CAt) model is limited in terms of predicting only Food, Staff and Ambience aspects, we make a slight change to the modelling setup to make it compatible to run on the single aspect sentences of all aspects. Please refer to the appendix for the details about this change. With the help of appropriate masks in the code, it is ensured that only the sentences corresponding to the three aspect categories for which prior model is good at predicting, are penalized for dissimilar predictions. We use the value of $\sigma$ as 0.1. All the other model hyperparameters were set as per recommendation by ABAE \cite{He2017} and CAt \cite{Tulkens2020} research papers. The best results for Rule Based and Regularization Based Ensemble are shown in the Table \ref{tab:my-table}.

For the experimentation setup used, no public benchmark is available. Since the goal of the research is to create an ensemble, we consider individual models as the baseline. In addition to these models we have the rule based model as another baseline to compare against since they typically perform better than the individual models. From the results in the table \ref{tab:my-table} it can be observed that both ensemble creation techniques have outperformed the individual baseline models. The regularization based ensemble performs better than the rule based ensemble on unseen data overall and also saves the user from the hassle of manually creating the ensemble rules and saves a lot of experimentation time. The performance of regularization based ensemble was also found to be more consistent and stable across all the experiments in comparison to the experiments for rule based ensemble.

\begin{table}[h]
\centering
\caption{Results for Rule Based and Regularization Based Ensembles}
\label{tab:my-table}
\resizebox{\columnwidth}{!}{%
\begin{tabular}{cccccc}
\toprule
\textbf{F1 Scores} & \textbf{CAt} & \textbf{ABAE} & \textbf{Rule Ensb} & \textbf{Reg Ensb} & \textbf{Support} \\
\midrule
\textbf{Food}               & 64.00          & 64.64          & 67.59              & \textbf{67.95}    & 498              \\
\textbf{Staff}              & 50.93          & 56.43          & 62.16              & \textbf{61.39}    & 224              \\
\textbf{Ambience}           & 33.06          & 58.43          & 53.89              & \textbf{63.44}    & 150              \\
\textbf{Price}              & NA             & 32.20          & 30.28              & \textbf{36.36}    & 98               \\
\textbf{Miscellaneous}      & NA             & 66.86          & 69.42              & \textbf{70.55}    & 526              \\
\midrule
\textbf{Macro Average}    & 49.33 & 55.71 & 56.67 & \textbf{59.94} & 1496 \\
\textbf{Weighted Average} & 32.25 & 61.44 & 63.60 & \textbf{65.36} & 1496 \\
\bottomrule
\end{tabular}%
}
\end{table}

\section{Conclusions}


We proposed an innovative method of ensemble creation where predictions from the prior model were injected in ABAE model loss function as a regularization term. We showed that this ensemble creation technique is capable of outperforming individual models and rule based ensembles. It also reduced experimentation time since no explicit rules had to be defined for regularization based ensemble creation. To our knowledge, this is the first research to create ensemble of models through regularization in the domain of unsupervised ABSA. 


\section{Future Work}

The regularization based ensemble can be tried with different prior models. The setup can be applied to other datasets. Models can be tuned specific to the datasets to obtain better results. Penalization through regularization can be regulated further with the help of class weights where an error made by the model on a relatively sparser class is penalized more. The regularization based ensemble method can also be used in combination with rule based ensemble creation where the rules can be applied later in the pipeline as a post processing step for the ensemble.

\appendix
\section{Additional Labels for CAt model}
Since the embeddings detected by the prior (CAt) model in this research are only for limited categories i.e. Food, Staff and Ambience, we need to provision for an additional category which would correspond to sentences belonging to none of the above categories. We need this aspect to be as far away as possible from the embeddings for the three aspects that the prior model predicts for. To do so we calculate the average of the embeddings of the three aspect categories and multiply the result by -1. In our case the closest match to this result was "baby". So we use the embedding of this as a place holder embedding for the additional aspect. In Regularization Based Ensemble, the CAt model thus predicts this additional category along with the other three aspects. 

\section{Additional Details about Experiments}
Some of the experiments done for the rule based ensemble are as follows:
\begin{enumerate}
    \item \textbf{NN-ADJ:} Use nouns and adjectives to find aspects for sentences as per similarity process where the two models don’t agree, but only for Food, Staff and Ambience. Use ABAE for the rest and miscellaneous to fill the ones not predicted by ABAE.
    \item \textbf{Only-NN:} Same as Experiment 1 with the exception of using only list of nouns instead of nouns and adjectives.
    \item \textbf{ABAE-misc:} Use only ABAE predictions and predict the rest as miscellaneous.
    \item \textbf{NN-ADJ-FoSt:} Same as Experiment 1 but, the disambiguation is only done for Food and Staff.
\end{enumerate}

Some of the experiments performed for regularization based ensemble are as follows:
\begin{enumerate}
    \item \textbf{Ensb-NN-ADJ:} For the predictions missed by the ensemble, the noun and adjective list is used for aspect assignment based on similarity, as done in rule-based ensemble.
    \item \textbf{Ensb-Only-NN:} Same as Experiment 1 but, here noun list- is used instead of the noun adjective list.
    \item \textbf{Ensb-misc:} Here, miscellaneous category is used to fill the predictions missed by the ensemble.
\end{enumerate}

For rule based ensemble, various experiments were performed where word similarity score was used in addition to the mutual predictions of the models, using noun list and noun adjective list. A general observation over unseen data revealed that, the ensemble of the models along with the rules performed better than individual models. Even the worst performing ensemble proved to be better than in individual models. In addition to this, limitation of performance of the CAt model was also overcome, which only works well for limited number of categories in the data. Since the model was used in combination with the ABAE model, not only was the CAt model utilized for full-scope, but better performance was obtained out of the ensembles on full-scope of the aspects. As per the research, the measurement of metrics even for ABAE model was done for the limited number of aspects. But using the ensemble, the predictions for all aspects were obtained and it has been shown that it is possible to use multiple models together even in the unsupervised setting, where there is no straight forward way of ensemble creation. By this it can be concluded that, it is possible to create ensemble of models in unsupervised setting with the help of embeddings and calculated rule creation in addition to mutual predictions, which can lead to better results than individual models. However, despite the better performance by rule based models, they were unable to perform better than the regularization based ensemble creation method.